\documentclass[runningheads]{llncs}

\usepackage{amsfonts}
\usepackage{amsmath}
\usepackage{graphicx}
\usepackage[utf8]{inputenc}
\usepackage{multirow}
\usepackage{pgfplots}
\usepackage{subcaption}
\usepackage{svg}

\DeclareMathOperator*{\argmin}{argmin}

\begin{document}

\title{Writer Identification and Writer Retrieval Based on NetVLAD with Re-ranking}

\titlerunning{WI and WR Based on NetVLAD with Re-ranking}

\author{Shervin Rasoulzadeh \and Bagher Babaali}

\authorrunning{S. Rasoulzadeh \and B. Babaali}

\institute{School of Mathematics, Statistics and Computer Science, College of Science, University of Tehran, Tehran, Iran\\
\email{\{s.rasoulzadeh,babaali\}@ut.ac.ir}}

\maketitle

\begin{abstract}
This paper addresses writer identification and writer retrieval which is considered as a challenging problem in the document analysis and recognition field. In this work, a novel pipeline is proposed for the problem at hand by employing a unified neural network architecture consisting of the ResNet-20 as a feature extractor and an integrated NetVLAD layer, inspired by the vector of locally aggregated descriptors (VLAD), in the head of the latter part. Having defined this architecture, the triplet semi-hard loss function is used to directly learn an embedding for individual input image patches. Subsequently, generalized max-pooling technique is employed for the aggregation of embedded descriptors of each handwritten image. Also, a novel re-ranking strategy is introduced for the task of identification and retrieval based on $k$-reciprocal nearest neighbors, and it is shown that the pipeline can benefit tremendously from this step. Experimental evaluation has been done on the three publicly available datasets: the ICDAR 2013, CVL, and KHATT datasets. Results indicate that while we perform comparably to the state-of-the-art on the KHATT, our writer identification and writer retrieval pipeline achieves superior performance on the ICDAR 2013 and CVL datasets in terms of mAP.
\end{abstract}

\section{Introduction}\label{sec:1_introduction}
Along with biometrics identifiers such as DNA, fingerprints, etc, handwriting is considered as a special case of behavioral biometrics \cite{schomaker2008writer}. Handwriting analysis helps to extract attributes such as writer from a handwritten document. Several factors may lead to handwriting variability such as having diseases (Parkinson could be recalled as an example), education, time and effort spent, different pens, etc \cite{christlein2019handwriting}. Thus, Handwriting analysis is considered a complex and challenging task. To overcome these challenges and to provide an automatic handwriting analysis system, one needs to differentiate between \textit{online} and \textit{offline} data/systems. Online text analysis systems capture the whole procedure of writing with special devices and the input consists of temporal data such as pen-tip positions. On the other hand, offline data is usually static and typically is in the format of an image. Also, methods for handwriting analysis are categorized into two categories: \textit{text-dependent} and \textit{text-independent} methods. In the text-dependent methods, each handwriting must contain a fixed content, while in the text-independent methods no assumptions are made on the content of handwriting and any arbitrary text could constitute the handwriting. In this work, it is intended to provide an offline text-independent handwriting analysis system concentrated on the problem of writer recognition, specifically writer identification and writer retrieval.

\begin{figure}
    \centering
    \includegraphics[scale=0.065]{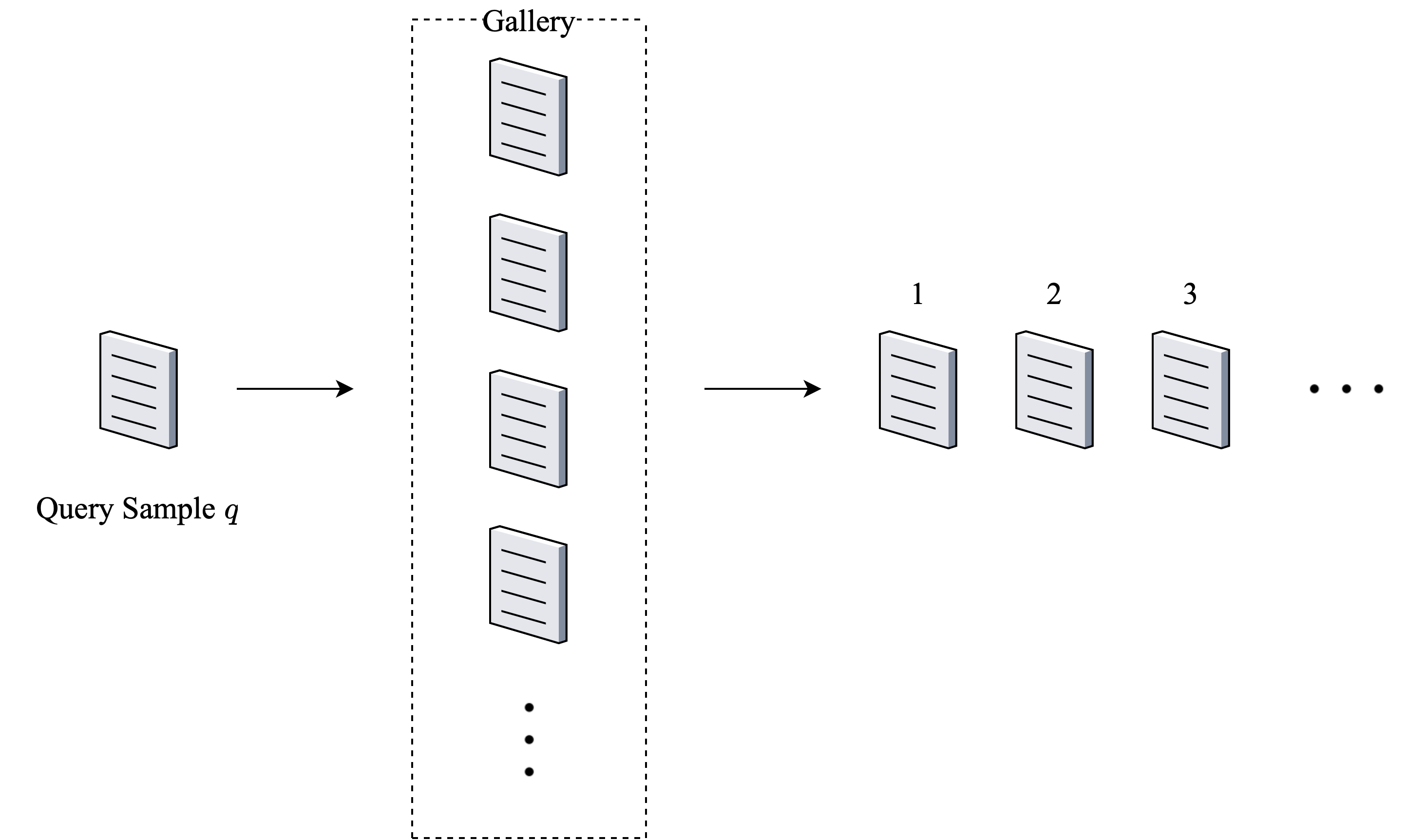}
    \caption{Overall schema of writer retrieval.}
    \label{fig:img_1}
\end{figure}

\begin{figure}
    \centering
    \includegraphics[scale=0.065]{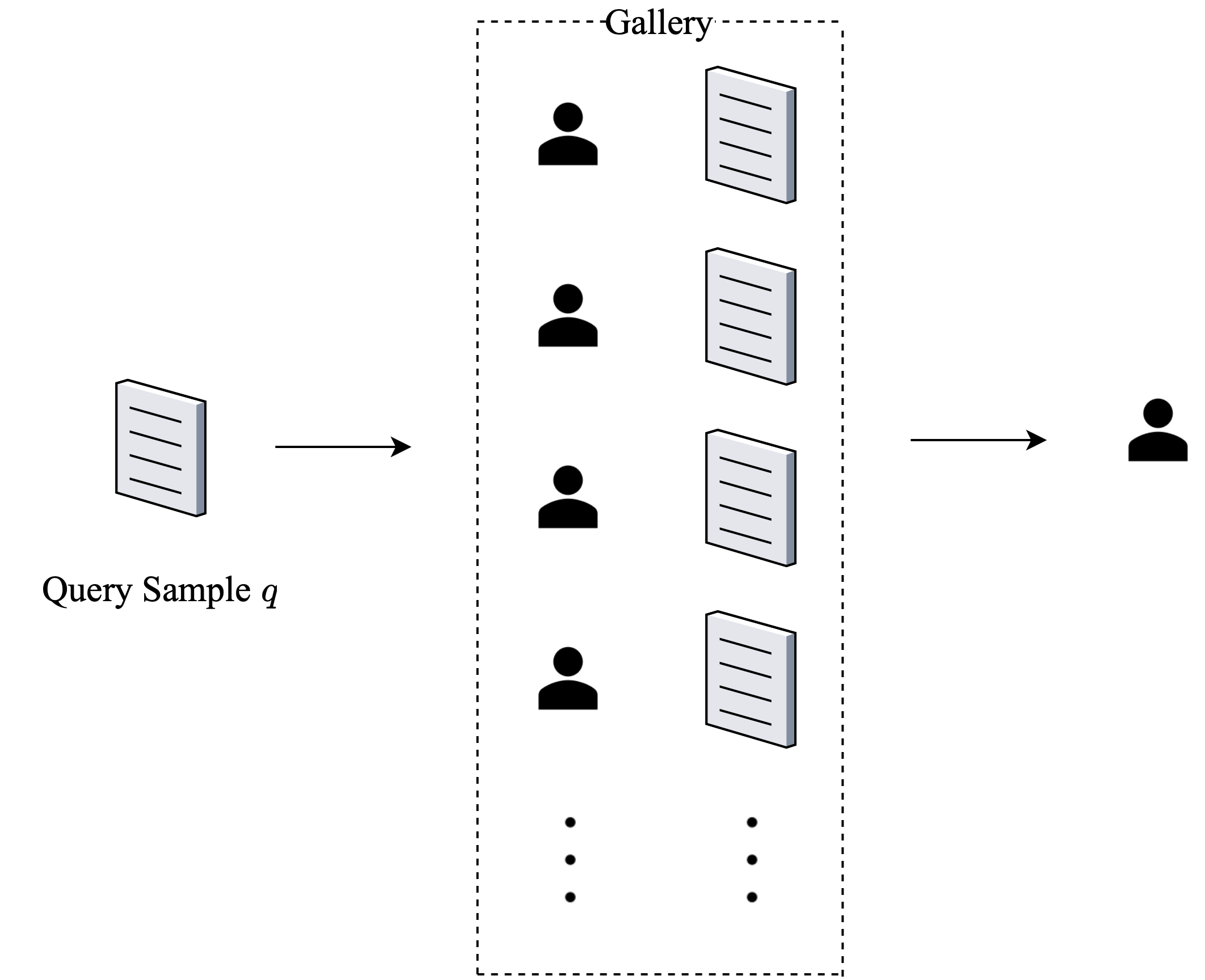}
    \caption{Overall schema of writer identification.}
    \label{fig:img_2}
\end{figure}

\indent \textit{Writer retrieval} is the task of ranking document images from a large corpus of data with similar handwriting to the \textit{query} sample, see Figure \ref{fig:img_1}. Experts in relative fields then analyze these rankings and thus new documents from the same writer can be found. Historians and paleographers benefit the most from this scenario. When analyzing historical documents, a vast amount of data should be dealt with, where examining them individually is not possible or a very time-consuming task. However, this scenario helps to find the writer of a historical document in a shortlists without having to go through all documents in the database. In contrast to writer retrieval, \textit{writer identification} is the task of finding the writer of a query sample assuming a set of documents where each one's writer is \textit{known} in advance, see Figure \ref{fig:img_2}. This scenario is often applicable in forensic sciences, e.g, finding the writer of a threatening letter.
The methods for the two scenarios are quite similar. Both generate a feature vector describing the handwriting of each document with respect to its writer. These feature vectors are compared using a similarity measure criterion such as cosine distance, $\chi_{2}$-distance, etc, and then accordingly rankings are computed. For the retrieval scenario, these rankings are further analyzed and a shortlist of documents written by the query's writer will be returned. On the other hand, for the identification scenario, the writer of the sample with the shortest distance to the query sample is returned as output.

\indent Our proposed pipeline uses a ResNet-20 convolutional neural network with the NetVLAD layer, inspired by Vectors of Locally Aggregated Descriptors (VLAD), to extract local descriptors and their corresponding embeddings describing small windows in a document image. Afterward, generalized max-pooling (GMP) aggregation is employed to achieve a single global image descriptor for each image from the computed embeddings. Dimensionality reduced global image descriptors resulted from applying the PCA, are then compared and rankings are computed. In the final stage, we make use of a re-ranking strategy based on query expansion and $k$-reciprocal nearest neighbors to improve the retrieved rankings before evaluation.

\indent The structure of this work is as follows: In Section \ref{sec:2_related_work} some deep-learning-based related work in the field of writer identification and retrieval, as well as re-rankings, are discussed. Afterwards, in Section \ref{sec:3_writer_identification_and_writer_retrieval_pipeline} we investigate our proposed pipeline in great detail. Section \ref{sec:4_evaluation} introduces datasets and finally, we evaluate and compare our proposed pipeline against state of the art on three different datasets (ICDAR 2013, CVL, and KHATT) in Section \ref{sec:5_conclusion}.

\section{Related Work}\label{sec:2_related_work}
Nearly all common writer identification and retrieval datasets consist of writer disjoint train and test sets. Hence, end-to-end training cannot be applied.\\
\indent One of the first writer recognition methods using deep learning techniques was proposed by Fiel and Sablatnig \cite{fiel2015writer}. They trained the "CaffeNet" CNN on the line and word segmentations. Feature vectors extracted from the penultimate layer of the CNN are compared using the $\chi^{2}$-distance. At the time, their results showed superior performance on the IAM \cite{marti2002iam} and ICFHR’12 datasets while being inferior on the ICDAR 2013 \cite{louloudis2013icdar} dataset. Christlein et al \cite{christlein2015offline} used activation features from a CNN as image local descriptors. Afterward, global image descriptors are formed through the GMM supervector encoding. Their approach improved 0.21 \% in terms of mAP on the ICDAR13 dataset. In another recent method by Christlein et al \cite{christlein2018encoding} LeNet and ResNet architectures are employed to extract local descriptors followed by VLAD encoding to compute global image descriptors for document images. They experimented with both exemplar support vector machines (ESVMs) and nearest neighbors to evaluate their pipeline. To the best of our knowledge, their approach has set new standards on the ICDAR 2013 and CVL \cite {kleber2013cvl} datasets.\\
\indent In \cite{jordan2020re} Jordan et al. experimented with reciprocal relationships in two ways. First, integrated them into the Jaccard distance and computed the final rankings based on a weighted sum of the original distance and the Jaccard distance. Second, encoded them into new feature vectors and hence expanded the positive set for ESVMs. As a result, both of their techniques outperformed the baseline on the ICDAR 2017 dataset \cite{fiel2017icdar2017}.\\
\indent Tang and Wu \cite{tang2016text} proposed a novel approach with convolutional neural network (CNN) and joint Bayesian consisting of two stages: 1. feature extraction and 2. writer identification.  They used CNNs to extract global features instead of small image patches. They used random word segmentations and generated 500 and 20 training samples per writer for training and testing, respectively. Finally, a Bayesian network is used for the computation of similarity between feature vectors. At the time, they achieved the best results compared to the state-of-the-art on ICDAR 2013 and CVL datasets. In another work by Xing and Qiao \cite{xing2016deepwriter}, two adjacent images patches used as inputs to their proposed network, named DeepWriter, consisting of two branches sharing the convolutional Layers. For the final evaluation part, two softmax layers belonging to each branch were averaged to predict the writer and achieved promising results on the IAM dataset. Despite that they worked in an end-to-end manner (i.e. "the CNN is trained for a specific writer on a line-basis using some lines for training, one for validation and one for testing." \cite{christlein2018encoding}.), comparison of their work with other pipelines in literature is impossible.\\
\indent Considering that, we could say that our proposed pipeline is mostly inspired by the works of Christlein et al \cite{christlein2015offline,christlein2018encoding} and Jordan et al. \cite{jordan2020re}. However, with our proposed pipeline consisting of the unified neural network architecture with the NetVLAD layer, and re-ranking strategy based query expansion and $k$-reciprocal nearest neighbors, we could improve upon the state of the art on the ICDAR 2013 and CVL datasets.

\section{The Proposed Pipeline for Writer Identification and Writer Retrieval}\label{sec:3_writer_identification_and_writer_retrieval_pipeline}
\begin{figure*}[t]
    \centering
    \includegraphics[scale=0.125]{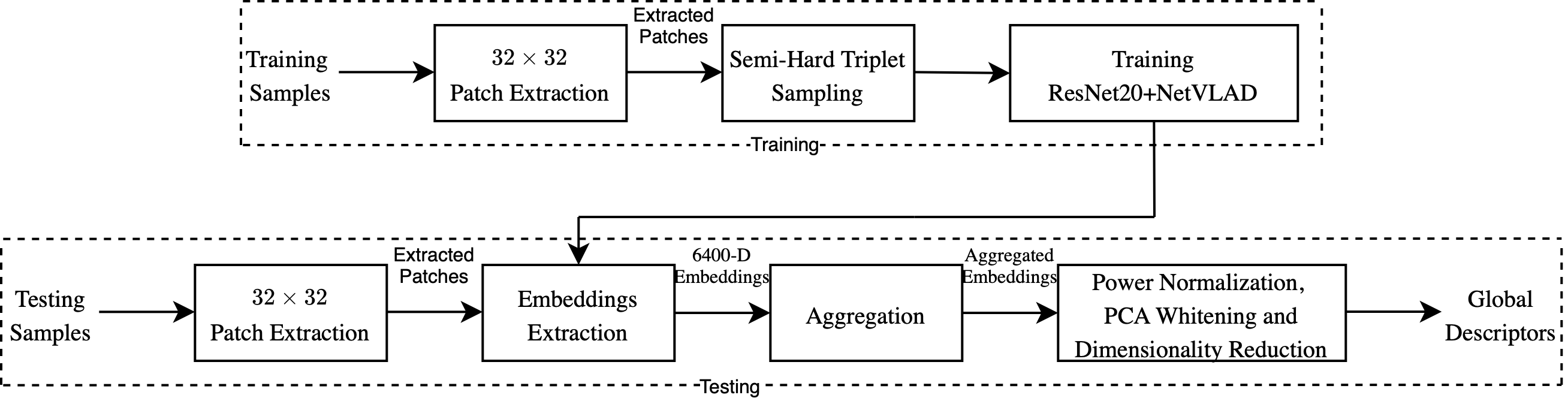}
    \caption{Overview of proposed pipeline.}
    \label{fig:img_3}
\end{figure*}

Our proposed pipeline consists of two parts: 1. A unified neural network architecture with ResNet-20 \cite{he2016deep,he2016identity} and the NetVLAD layer \cite{arandjelovic2016netvlad}, and 2. A re-ranking strategy to improve the final results. The first part itself consists of three main steps (depicted in Figure \ref{fig:img_3}): The ResNet-20 with the NetVLAD layer to extract the local image descriptors and their corresponding embeddings, An orderless aggregation function to the pool obtained embeddings of each image into one global image descriptor, and the normalization and PCA \cite{wold1987principal} based dimensionality reduction of the resulted global image descriptors.

\subsection{Convolutional Nueral Network with NetVLAD Layer}\label{sub_sec:convolutional_neural_network_with_netvlad_layer}
State-of-the-art deep-learning-based methods in writer identification and writer retrieval usually employ a CNN to extract the local image descriptors which are subsequently encoded using an encoding method. An encoding consists of two steps: 1) An \textit{embedding} step, where local feature vectors are projected into a possibly high-dimensional space, and 2) An \textit{aggregation} step, in which embedded local feature vectors of each input image are pooled into one global image descriptor. Christlein et al. \cite{christlein2018encoding} computed the local feature vectors through ResNet-20 residual neural network and used the VLAD \cite{jegou2011aggregating} encoding method for embedding and aggregation. Building on the success of Christlein et al. \cite{christlein2018encoding} we propose a unified Neural network consisting of ResNet-20 followed by the trainable NetVLAD layer \cite{arandjelovic2016netvlad}, inspired by the VLAD, at the head of the last convolutional layer of ResNet-20 to learn the embedding of feature vectors in an end-to-end manner using a triplet loss \cite{schroff2015facenet}.

\subsubsection{ResNet-20 and NetVLAD}\label{sub_sub_sec:resnet20_and_netvlad}
\begin{center}
    \begin{figure}[t]
        \centering
        \includegraphics[scale=0.30]{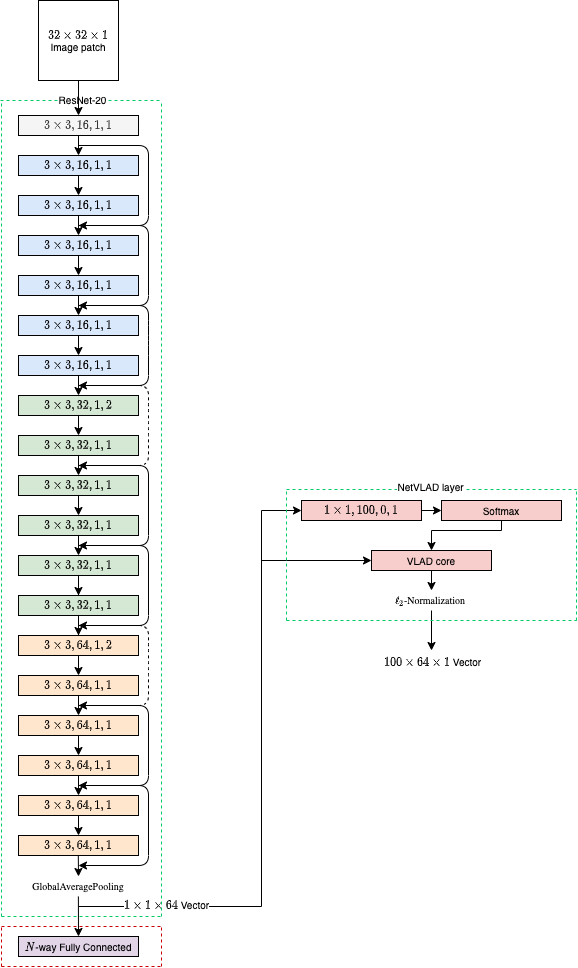}
        \caption{CNN architecture composed of the ResNet-20 followed by the NetVLAD layer. Numbers in each rectangle denote \textit{kernel size}, \textit{number of output filters,} \textit{padding}, and \textit{size of stride}, respectively. The $N$-way fully connected is dropped and instead the $1 \times 1 \times 64$ output vector is passed to the NetVLAD layer.}
        \label{fig:img_4}
    \end{figure}
\end{center}

Details of the ResNet-20 convolutional neural network and the NetVLAD layer are described in the following.
\paragraph*{ResNet-20 Convolutional Neural Network.} For network inputs, $32 \times 32$ image patches centered at the contour of handwriting were extracted. Same as Christlein et al. \cite{christlein2018encoding} we follow the architectural design of He et al. \cite{he2016deep} on CIFAR10 dataset \cite{krizhevsky2009learning}. $6n + 2$ layers are employed with $n$ set to $3$ leading to the ResNet-20 architecture. The first layer is $3 \times 3$ convolutions. Then an stack of $6n$ layers with $3 \times 3$ convolutions follows with every $2n$ layers forming an \textit{stage}. At the beginning of each stage (except the first one), the feature map size is halved (downsampled) by a convolutional layer with the stride of $2$, while the number of filters is doubled. Within each stage, the layers have the same number of filters. More precisely, feature maps and filters for stages are of sizes $\lbrace 32, 16, 8 \rbrace$ and $\lbrace 16, 32, 64 \rbrace$, respectively. Shortcut connections are connected to the pairs of $3 \times 3$ layers leading to a total $3n$ shortcuts. The network ends with the global average pooling layer with a size of $8$ and an $N$-way fully connected layer. However, We discard the last fully-connected layer and pass the $1 \times 1 \times 64$ output feature vector of global average pooling layer to the NetVLAD to further learn the VLAD embeddings of these feature vectors. (See Figure \ref{fig:img_4}).
\paragraph*{NetVLAD Layer.} The idea behind the vectors of locally aggregated descriptors (VLAD) \cite{jegou2011aggregating} is to compute the embeddings by means of residuals $x_{i} - c_{j}$ for each local image descriptor $x_{i}$. Finally, embedded local image descriptors of each image are accumulated by an orderless aggregation function. This characterizes the distribution
of the vectors with respect to the cluster centers. The VLAD embedding can be regarded as a simplified version of the Fisher Vectors \cite{jegou2011aggregating}. More precisely, given $N$ local image descriptors $\lbrace x_{i} | x_{i} \in \mathbb{R}^{D}, i=1, \dots, N \rbrace$ and a dictionary of $K$ cluster centers $\lbrace c_{j} | c_{j} \in \mathbb{R}^{D}, j=1, \dots, K\rbrace$, the VLAD embedding function is computed as follows:
\begin{equation}
    \label{eq:eq_1}
    \phi_{\text{VLAD}, k}(x_{i}) = \alpha_{k}(x_{i})(x_{i} - c_{k})
\end{equation}
\begin{equation}
    \label{eq:eq_2}
    \alpha_{k}(x_{i}) = \begin{cases}
     1, & \text{if } k = \argmin_{j=1, \dots K} \Vert x_{i}-c_{j} \Vert\\
    0, & \text{else}
    \end{cases}
\end{equation}
Then $\phi_{\text{VLAD}}(x_{i}) = (\phi_{1}(x_{i}), \dots, \phi_{K}(x_{i}))$ represents the full embedding for each local image descriptor $x_{i}$.\\
\indent Arandjelovic et al \cite{arandjelovic2016netvlad} introduced a trainable generalized VLAD layer, named NetVLAD, which is pluggable into any CNN architecture. Clusters hard-assignments $a_k(x_{i})$ of local image descriptors in the original VLAD are the source of discontinuities which prevent differentiability in the backpropagation procedure. The authors replaced them with soft-assignment to make it amenable to backpropagation:
\begin{equation}
    \label{eq:eq_3}
    \Bar{\alpha}_{k}(x_{i}) = \frac{e^{-\alpha\Vert x_{i}-c_{k} \Vert^{2}}}{\sum_{j=1}^{K}e^{-\alpha\Vert x_{i}-c_{j} \Vert^{2}}}
\end{equation}
where $\alpha$ is a parameter that control the decay of response with the magnitude of distance. Intuitively, Equation (\ref{eq:eq_3}) assigns the weights of local image descriptors $x_{i}$ proportional to their nearness to clusters $c_{j}$. Moreover, factorizing $e^{-\alpha\Vert x_{i} \Vert^{2}}$ results in:
\begin{equation}
    \label{eq:eq_4}
    \Bar{\alpha}_{k}(x_{i}) = \frac{e^{w_{k}^{T}x_{i}+b_{k}}}{\sum_{j=1}^{K} e^{w_{j}^{T}x_{i}+b_{j}}}
\end{equation}
where $w_{k} = 2\alpha c_{k}$ and $b_{k} = -\alpha\Vert c_{k} \Vert^{2}$. However, in \cite{arandjelovic2013all} the authors propose decoupling dependencies of parameters $c_{k}$, $w_{k}$, and $b_{k}$ as it will brings greater flexibility to the model. In this manner, NetVLAD layer consists of three independent set of learnable parameters. We crop the ResNet-20 at the last convolutional layer and view it as a $D$-dimensional (here $D = 64$) local image descriptor. As depicted in Figure \ref{fig:img_4}, the NetVLAD layer can be decomposed into CNN layers connected in an acyclic graph. Equation (\ref{eq:eq_4}) represents the soft-max activation function. So the soft-assignments of local image descriptor $x_{i}$ to clusters $c_{k}$ can be viewed as applying a $1 \times 1$ convolution layer with $K$ filters representing $w_{k}$ and biases as $b_{k}$ followed by the soft-max activation function to obtain final soft-assignments $\Bar{\alpha}_{k}(x_{i})$. The final output is $K \times D \times 1$-dimensional representing the full embedding for local image descriptor $x_{i}$.\\
The authors in \cite{arandjelovic2013all} regard the output of the last convolutional layer with $H \times W \times D$ map as a set of $D$-dimensional descriptors at $H \times W$ spatial locations in input image which are further embedded and pooled by the NetVLAD layer. 
However, by using ResNet-20  with image patches of size $32 \times 32$ as feature extractor, the output of the last convolutional layer becomes $1 \times 1 \times 64$ map which we consider as $64$-dimensional local image descriptor extracted from the input image patch. Passing this descriptor (with $H =1$ and $W = 1$) enables NetVLAD layer to learn the respective local image descriptor embedding. So in this manner, the NetVLAD layer functions to learn the generalized VLAD embeddings.  Prior to forwarding the embeddings extracted from the NetVLAD layer to the triplet loss, they are $\ell_{2}$-normalized.
\paragraph*{Learning from Semi-Hard Triplets.} A quite well strategy to learn the parameters of the network to reach a good encoding is through triplet loss function. We wish to learn VLAD embedding representation $\phi_{\text{VLAD}}(x)$ constrained to lie on $K \times D$-dimensional hypersphare, i.e. $\Vert \phi_{\text{VLAD}}(x) \Vert = 1$, such that two embeddings belonging to the image(s) of the same writer be close together in the embedding space while embeddings of images with different writers lie far away from each other. The only requirement is that given two positive embeddings of the same writer and one negative embedding, the negative should be farther away than the positive by some margin $m$. This requirement can be translated into a loss between triplets. The loss will be defined over triplets of embeddings: an anchor $\phi_{a}$, a positive of the same writer as the anchor $\phi_{p}$, and a negative of a different writer $\phi_{n}$. For some distance on the embedding space $d$, the loss of a triplets $(\phi_{a}, \phi_{p}, \phi_{n})$ is:
\begin{equation}
    \label{eq:eq_5}
    \mathcal{L} = \max (d(\phi_{a}, \phi_{p}) - d(\phi_{a}, \phi_{n}) + margin, 0)
\end{equation}
\indent The original NetVLAD paper utilizes the weakly supervised triplet ranking loss \cite{arandjelovic2016netvlad}. However, Since here the NetVLAD layer is applied to learn mbeddings in a patch-wise manner, another strategy is employed. Based on the definition of loss we tend to train on semi-hard triplets \cite{schroff2015facenet}: triplets where the negative is not closer to the anchor than the positive, but which still produce positive loss: 
\begin{equation}
    \label{eq:eq_5_1}
    d(\phi_{a}, \phi_{p}) < d(\phi_{a}, \phi_{n}) < d(\phi_{a}, \phi_{p}) + margin.
\end{equation}
We train parameters of the proposed pipeline on a large set of semi-hard triplets image patches triplets extracted from the respective dataset. Details and parameters of training are given in Section \ref{sec:4_evaluation}.

\subsubsection{Aggregation}\label{sub_sub_sec:aggregation}
Aggregation step is required to obtain a single vector representing each image from its embedded local descriptors. Default aggregation method is \textit{sum-pooling}. Assuming the set of $N$ local descriptors $\mathcal{X} = \lbrace x_{i} | i = 1, \cdots, N \rbrace$ for an image, sum-pooling constructs global descriptor $\xi$ as follows:
\begin{equation}
    \label{eq:eq_6}
	\xi = \psi(\phi(\mathcal{X}))	 = \sum_{x \in \mathcal{X}} \phi(x).
\end{equation}
"Since we sum over all descriptors, the aggregated descriptors can suffer from interference of unrelated descriptors that influence the similarity, even if they have low individual similarity" \cite{christlein2019deep} as the similarity $\mathcal{K}(\mathcal{X}, \mathcal{Y})$ between two images represented by sets $\mathcal{X}$ and $\mathcal{Y}$ is computed as follows:
\begin{equation}
    \label{eq:eq_7}
	\mathcal{K}(\mathcal{X}, \mathcal{Y}) = \sum_{x \in \mathcal{X}} \sum_{y \in \mathcal{Y}} \phi(x) \cdot \phi(y).
\end{equation}
Hence, more frequently occurring descriptors will be more influential in the final representation and affect the final similarity between global descriptors. This phenomenon is called \textit{visual burstiness} \cite{jegou2009burstiness}. Recently, a novel approach named \textit{generalized max-pooling} \cite{murray2014generalized} was proposed to overcome this problem and has successfully applied in the field of writer identification and writer retrieval in works of Christlein et al. \cite{christlein2018encoding}. We employed this method in our pipeline as it has shown superior performance to the other two methods \cite{murray2016interferences}. Generalized max-pooling balances contribution of every embedding $\phi(x) \in \mathbb{R}^{K \times D}$ where $x \in \mathbb{R}^{D}$ is local image descriptor, by solving a ridge regression problem. Therefore, 
\begin{equation}
    \label{eq:eq_8}
    \phi(x)^{T} \xi_{gmp}(\mathcal{X}) = C, \quad \forall x \in \mathcal{X},
\end{equation}
where $\mathcal{X}$ is the set of all local descriptors of an image, $\xi_{gmp}$ denotes aggregated global image descriptor and $C$ is a constant that can be set arbitrarily since it does not influence the global image descriptors sine they are subsequently normalized in the post-processing step. Equation (\ref{eq:eq_8}) can be re-formulated for all $N$ local image descriptors of each image as below:
\begin{equation}
    \label{eq:eq_9}
    \Phi^{T}\xi_{gmp} = 1_{N},
\end{equation}
where $\Phi$ and $1_{N}$ denote the $(K \times D) \times N$ matrix of all local image descriptors embeddings and vector of $N$ constants set to $1$, respectively. Equation (\ref{eq:eq_9}) can be turned into a least-squares ridge regression problem \cite{christlein2018encoding,christlein2019handwriting}:
\begin{equation}
    \label{eq:eq_10}
    \xi_{gmp} = \argmin_{\xi}||\Phi^{T}\xi - 1_{N}||_{2}^{2} + \lambda||\xi||_{2}^{2}
\end{equation}
with $\lambda$ being a regularization parameter. In the remainder of this work, $\psi$ denotes the aggregated global image descriptor.

\subsubsection{Normalization and Dimensionality Reduction}\label{sub_sec:normalization_and_dimensionality_reduction}
While working with global image descriptors obtained in the previous step, two challenges arise: (1) visual burstiness might corrupt visual similarity measure between the global image descriptors, i.e. the cosine distance used to rank images, and (2). These global descriptors lie in a very high-dimensional space and pipeline might benefit from projecting them to a lower-dimensional space. We address these challenges with an additional normalization and dimensionality reduction step.
\paragraph*{Power Normalization.} A normalization method to counter visual burstiness is named \textit{power normalization} \cite{perronnin2010improving} that proposes to apply function $f$ component-wise to global image descriptor $\psi$, 
\begin{equation}
    \label{eq:eq_11}
    f(\psi) = sign(\psi_{i})|\psi_{i}|^{p},\quad\forall i,1\leq i\leq n
\end{equation}
where $p$ is a normalization parameter and is generally is set to $0.5$. Power normalization is followed by $\ell_{2}$-normalization.
\paragraph*{Principal Component Analysis.} Due to the nature of VLAD encoding, global image descriptors lie in a very high-dimensional space. Principal component analysis (PCA) \cite{wold1987principal} is used to dimensionality reduce the encoding representations. However, this introduces a new parameter, $\textit{dimension}$, to the pipeline denoting the number of components to keep. After performing the PCA, $\ell_{2}$-normalization along each sample is necessary.

\subsection{Re-ranking}\label{sub_sec:re_ranking}
Writer identification and retrieval systems are evaluated using leave-one-image-out cross validation. Each image is once used as \textit{query} $q$ and the pipeline returns a ranked list $L(q)$ of all other images in test set (a.k.a \textit{gallery}). These ranked lists are obtained by computing the pairwise distance between query $q$ and each $p \in L(q)$ using a similarity measure criterion, i.e. \textit{cosine distance}. Given two vectors $p$ a $q$, the cosine distance is defined as:
\begin{equation}
    \label{eq:eq_12}
    d_{\text{cos}} (p, q) = 1 - \frac{pq}{\Vert p \Vert \Vert q \Vert}.
\end{equation}
Our goal is to re-rank each $L(q)$ based on knowledge lied in it, so that more \textit{relevant} samples rank top in the list and thus, boost the performance of writer identification and retrieval.

\subsubsection{Nearest Neighbors}\label{sub_sub_sec:nearest_neighbors}
$k$-nearest neighbors $k\text{NN}(q)$ (top-$k$ ranked samples of ranked list) of query $q$ is defined as:
\begin{equation}
    \label{eq:eq_13}
    k\text{NN}(q) = \lbrace p_{1}, p_{2}, \cdots, p_{k} \rbrace, \quad |k\text{NN}(q)| = k,
\end{equation}
Where $|.|$ denotes the cardinality of the set. The $k$-reciprocal nearest neighbors $k\text{rNN}(q, k)$ is defined as:
\begin{equation}
    \label{eq:eq_14}
    k\text{rNN}(q) = \lbrace p_{i} | p_{i} \in k\text{NN}(q) \land q \in k\text{NN}(p_{i}) \rbrace.
\end{equation}
In other words, two samples $q$ and $p$ are considered as $k$-reciprocal nearest neighbors, when both appear within the top-$k$ ranked samples of each other. According to the previous descriptions, $k$-reciprocal nearest neighbors are more related to query $q$ than $k$-nearest
neighbors.

\subsubsection{Query Expansion and $k$-Reciporal Nearest Neighbors}\label{sub_sub_sec:query_expansion_and_k_reciporal_nearest_neighbors}
A common approach to boost the performance of the information retrieval systems is automatic query expansion (QE) \cite{jordan2020re}. With an initial ranked list $L(q)$ computed, query expansion reformulates each query sample $q$ and obtains the improved ranked list by re-quering using the newly formed query instead of $q$.\\
\indent Chum et al \cite{chum2007total} proposed the following query expansion approach. For query $q$, a new query sample can be formed by taking the average over top-$n$ spatially verified samples $\mathcal{F}$ from ranked list $L(q)$,
\begin{equation}
    \label{eq:15}
    q_{\text{avg}} = \frac{1}{|\mathcal{F}|+1} \left(q + \sum_{f \ in \mathcal{F}} f \right)
\end{equation}
where $f_{i}$ and $n$ denote the $i$th sample in $\mathcal{F}$ and total number of samples in $|\mathcal{F}|$, respectively.\\
In our problem, the features do not encode any global spatial information and thus, we have no spatial verification at hand. Averaging over top-$k$ samples in $L(q)$ is not much reliable since the top-$k$ samples might contain false matches. We propose to use a more constrained strategy by taking an average over query $q$ and its $k$-rNNs in the initial ranked list to minimizing the risk of including false matches. Hence, the newly formed query $q_{\text{new}}$ is computed as follows: 
\begin{equation}
    \label{eq:eq_16}
	q_{\text{new}} = \frac{1}{|k\text{rNN}(q)| + 1} \left(q + \sum_{r \in k\text{rNN}(q)} r \right).
\end{equation}
This however introduces a new hyper-parameter $k$ to the pipeline. In the following, our proposed pipeline(with re-ranking) is denoted as "$\text{Proposed(+krNN-QE}_{k})$".

\section{Evaluation}\label{sec:4_evaluation}

\subsection{Datasets}\label{sub_sec:datasets}
\begin{figure}[tp]
    \centering
    \includegraphics[scale=0.35]{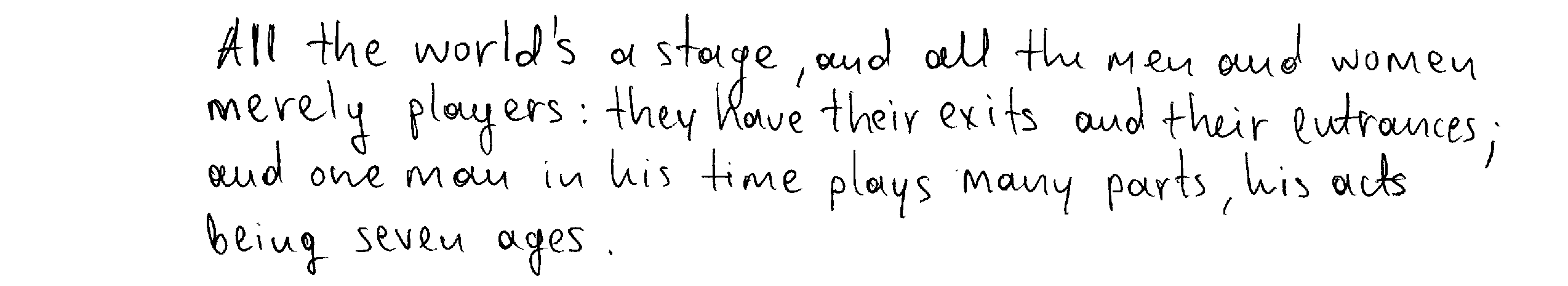}
    \caption{Example document from the ICDAR13 dataset (ID:032\_1)}
    \label{fig:img_5}
\end{figure}

\begin{figure}[tp]
    \centering
    \includegraphics[scale=0.095]{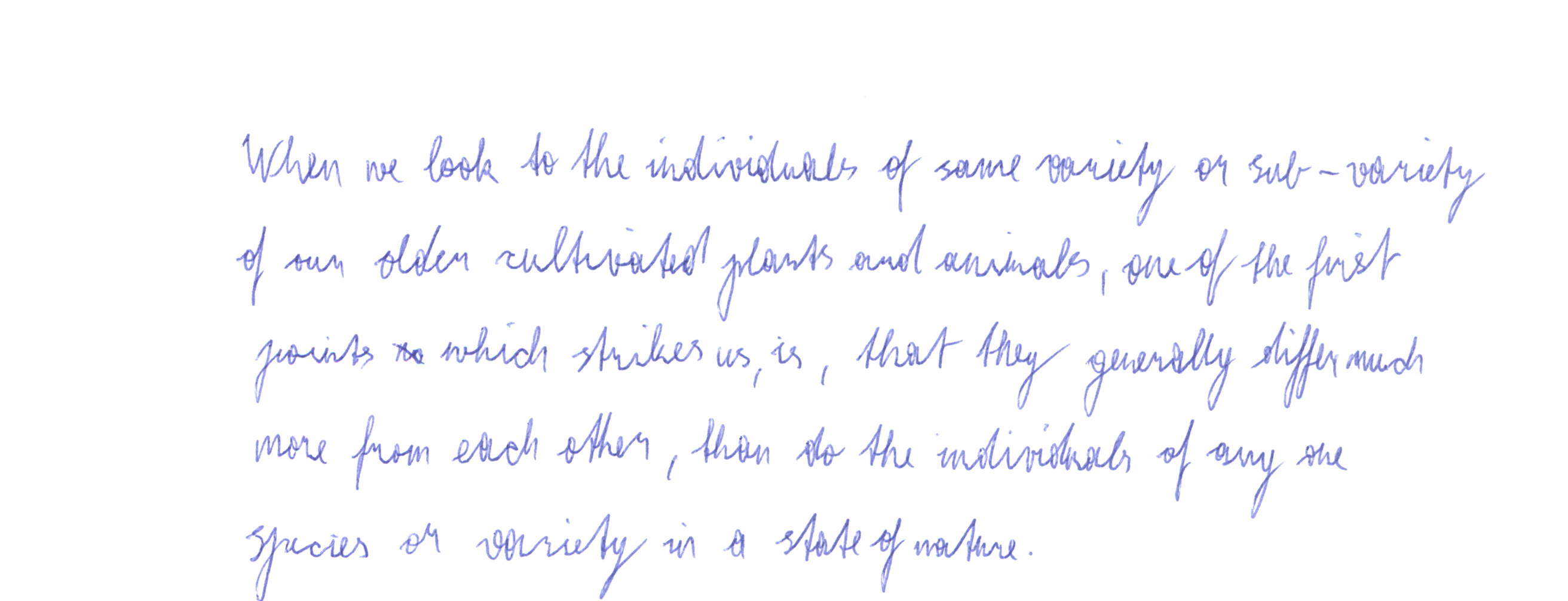}
    \caption{Example document from the CVL dataset (ID:0073-4).}
    \label{fig:img_6}
\end{figure}

\begin{figure}[tp]
    \centering
    \includegraphics[scale=0.45]{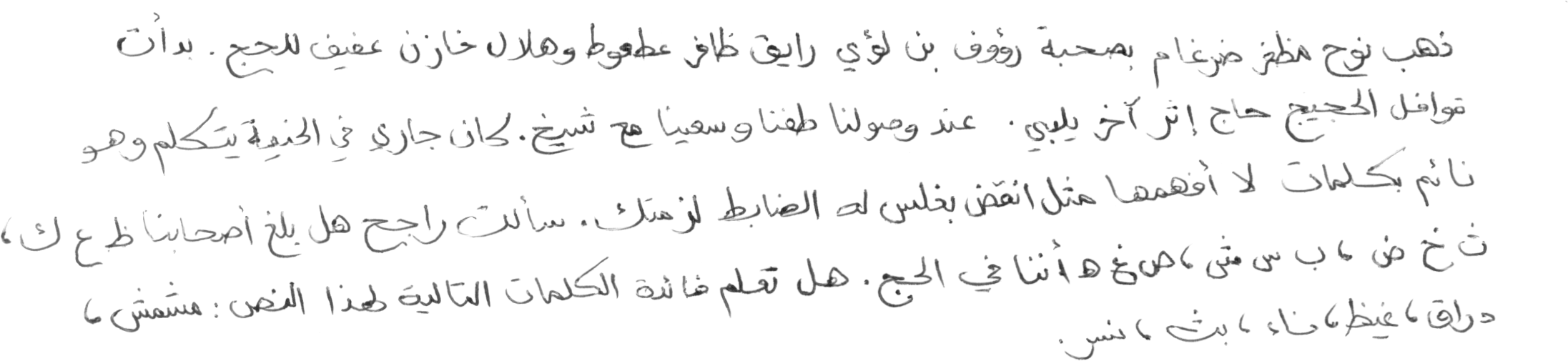}
    \caption{Example document from the KHATT dataset (ID:AHTD3A0003-1).}
    \label{fig:img_7}
\end{figure}

Our primary dataset is the ICDAR 2013. However, we compare our results against the state-of-the-art on the CVL and KHATT datasets as well. These datasets contain documents written in English, Greek, German, and Arabic. Successful results on these datasets also show that the proposed pipeline is language independent.
\paragraph*{ICDAR 2013}: The ICDAR 2013 dataset was introduced for the ICDAR 2013 competition on writer identification \cite{louloudis2013icdar}. It consists of four samples per writer, two of which are written in English, while the two others are in Greek. This dataset contains two disjoint train (a.k.a experimental) and test (a.k.a benchmarking) subsets. The train set consists of 100 writers, while the test set contains 250 writers. An example document from ICDAR 2013 dataset can be seen in Figure \ref{fig:img_5}.
\paragraph*{CVL}: The CVL dataset \cite{kleber2013cvl} (version 1.1) consists of 27 writers contributing to seven texts (one in German and six in English) in the officially provided train set. The test set consists of 283 writers, where each writer has copied five texts (one in German and four in English). Figure \ref{fig:img_6} depicts one sample document from this dataset.
\paragraph*{KHATT}: The KHATT dataset \cite{mahmoud2014khatt} used as a database for ICFHR 2014 \cite{slimane2014icfhr2014} Arabic writer identification competition. It consists of 1000 writers contributing to four samples. Two of which contain similar texts, while the other two were unique to each writer. The dataset is provided by three disjoint, i.e. writer independent sets, with a 4: 1: 1 ratio for train, test, and validation subsets, respectively. Along with two other datasets, a sample document image of this dataset is shown in Figure \ref{fig:img_7}.

\subsection{Metrics}\label{sub_sec:metrics}
Results reported in terms of the Hard Top-$N$ and mAP which are quite common metrics in information retrieval tasks. These metrics defined in the following.
\paragraph*{Hard Top-$\boldsymbol{N}$,} The strictest evaluation metric is hard top-$N$. A returned list $L(q)$ for query sample $q$ is considered as \textit{acceptable} \cite{christlein2014writer} if \textit{all} of the top-$N$ ranked samples in $L(q)$ belong to the same class as sample $q$'s class i.e., written by same writer. The ratio of the number of acceptable returned lists and the number of query samples is reported as Hard Top-$N$ accuracy.
\paragraph*{Mean Average Precision.} Another commonly used measure to evaluate an information retrieval task is the mean average precision (mAP) which considers the ranking of correct samples. It is calculated as the mean over all examined query samples $q$ of set $\mathcal{Q}$:
\begin{equation}
\label{eq:eq_17}
    \text{mAP} = \frac{\sum_{q \in \mathcal{Q}}\text{AveP}(q)}{|\mathcal{Q}|},
\end{equation}
where $\text{AveP}(q)$ is the average precision for a given query $q$ defined as below:
\begin{equation}
\label{eq:eq_18}
    \text{AveP}(q) = \frac{\sum_{k = 1}^{n}(P(k) \times rel(k))}{\text{number of relevant documents}},
\end{equation}
where $n$ is the total number of retrieved samples, $rel(k)$ is a binary function returning $1$ if sample at rank $k$ of $L(q)$ is relevant and $0$ otherwise, and $P(k)$ is the precision at rank $k$ (fraction of relevant items up to first $k$ retrieved samples in $L(q)$).

\subsection{Experiments and Results}\label{sub_sec:experiments_and_results}
Most datasets in the field of writer identification and writer retrieval come with disjoint train and test subsets and because of that, an end-to-end procedure would not be applicable. Therefore, our pipeline is composed of two phases: (1) training phase and (2) testing phase, each described below.
\paragraph*{Training phase.} To train and validate the neural network, $32 \times 32$ patches centered on the contour of handwritten images in the ICDAR 2013 train set get extracted. We sample around $5000000/25000$ image patches for train/validation where they are subsequently passed forward to the network. ResNet weights are initialized by \textit{He-initialization} \cite{he2015delving} and \textit{Xavier-initilization} \cite{glorot2010understanding} used to initialize Conv layer of the NetVLAD. As the ICDAR 2013 train set consists of $100$ writers, the number of cluster centers has been set to $100$ in the NetVLAD layer. The proposed neural network is optimized using Adamax with respect to triplet semi-hard loss with margin $m =0.1$, decay rates $\beta_{1} = 0.9$ and $\beta_{2} = 0.99$ for 1st moment estimate and exponentially weighted infinity norm, respectively. Training is stopped after $5$ epochs since the loss value stagnated at this point. The learning curves visualization is depicted in Figure \ref{fig:img_8}.
\paragraph*{Testing Phase.} Once the proposed network trained, we pass image patches of the test set of the dataset under evaluation to obtain embedded feature vectors where they are eventually pooled to compute global image descriptor of each image. For generalized max-pooling, following the works of Christlein et al. \cite{christlein2018encoding} we set $\lambda = 1000$. The next step after normalization ($\ell_{2}$ and SSR applied in order) is dimensionality reduction through PCA. PCA used for whitening and projecting global descriptors to a user-defined number of \textit{$dimension$}s (Figure \ref{fig:img_9} shows the mAP for the different number of \textit{$dimension$}s on ICDAR 2013 train and test subsets) after which they subsequently get $\ell_{2}$-normalization. The final step is following the evaluation procedure, i.e (re-)ranking documents based on their similarity. To have a clear view of how the proposed pipeline can benefit from re-ranking, Table \ref{tab:table_1} provides a comparison between the proposed pipeline with initial rankings (denoted as "$\text{ResNet-20+NetVLAD+NN}$") and re-ranking with different values for $k$ (denoted as "$\text{ResNet-20+NetVLAD+krNN-QE}_{k}$") on the ICDAR 2013 test set. As observations reveal that $\textit{dimension} = 128$ and $k = 2$ give the best results on the test set of this dataset, we follow our evaluation with these two parameters set.
\begin{figure}[t]
	\captionsetup[sub]{justification=centering}
	\centering
	\begin{subfigure}[t]{0.45\textwidth}
		\centering
		 \includegraphics[width=\textwidth, scale=0.5]{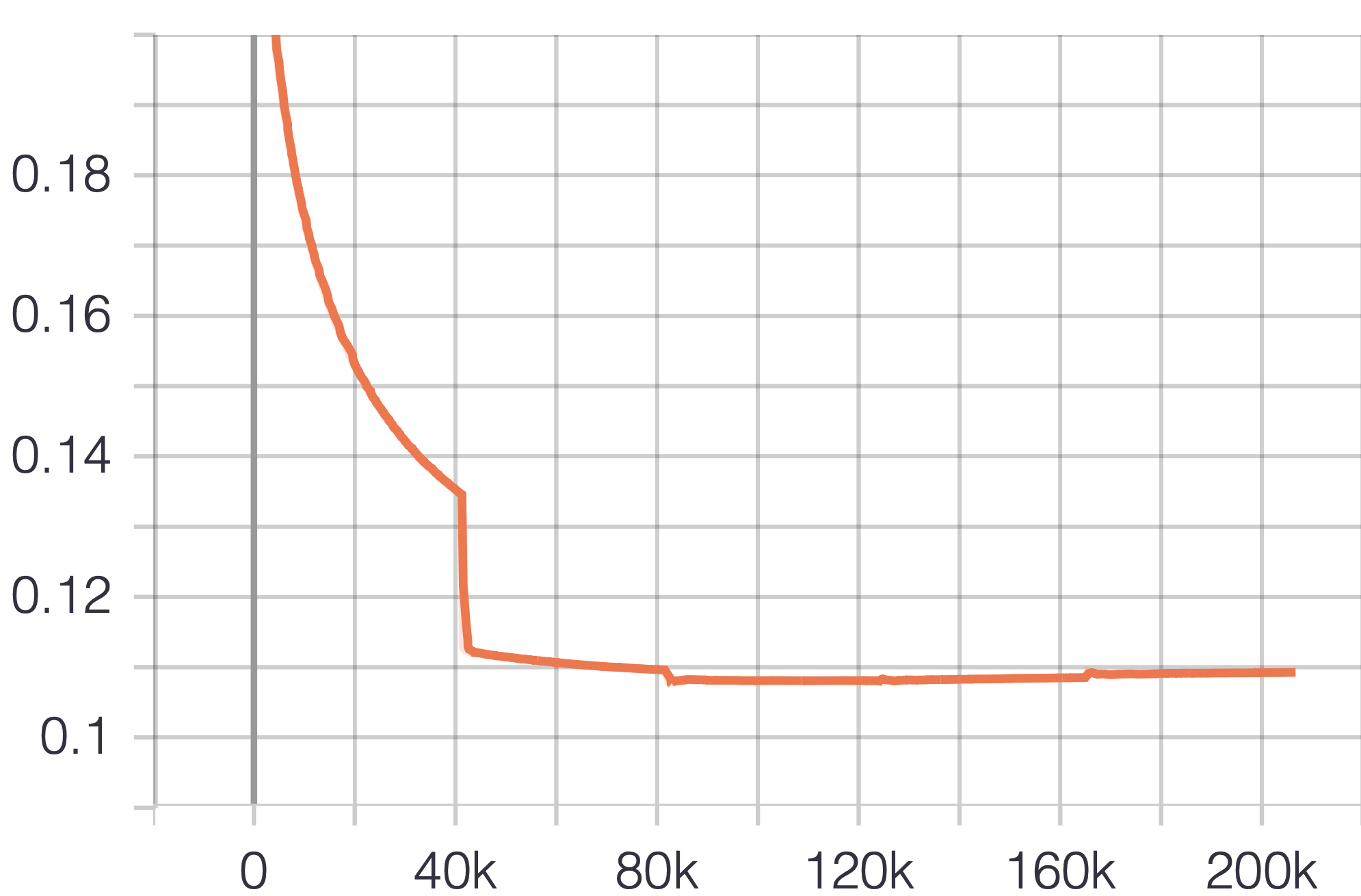}
		 \caption[]{}
		 \label{fig:img_8_1}
	\end{subfigure}
	\quad
	\begin{subfigure}[t]{0.45\textwidth}
		\centering
		\includegraphics[width=\textwidth, scale=0.5]{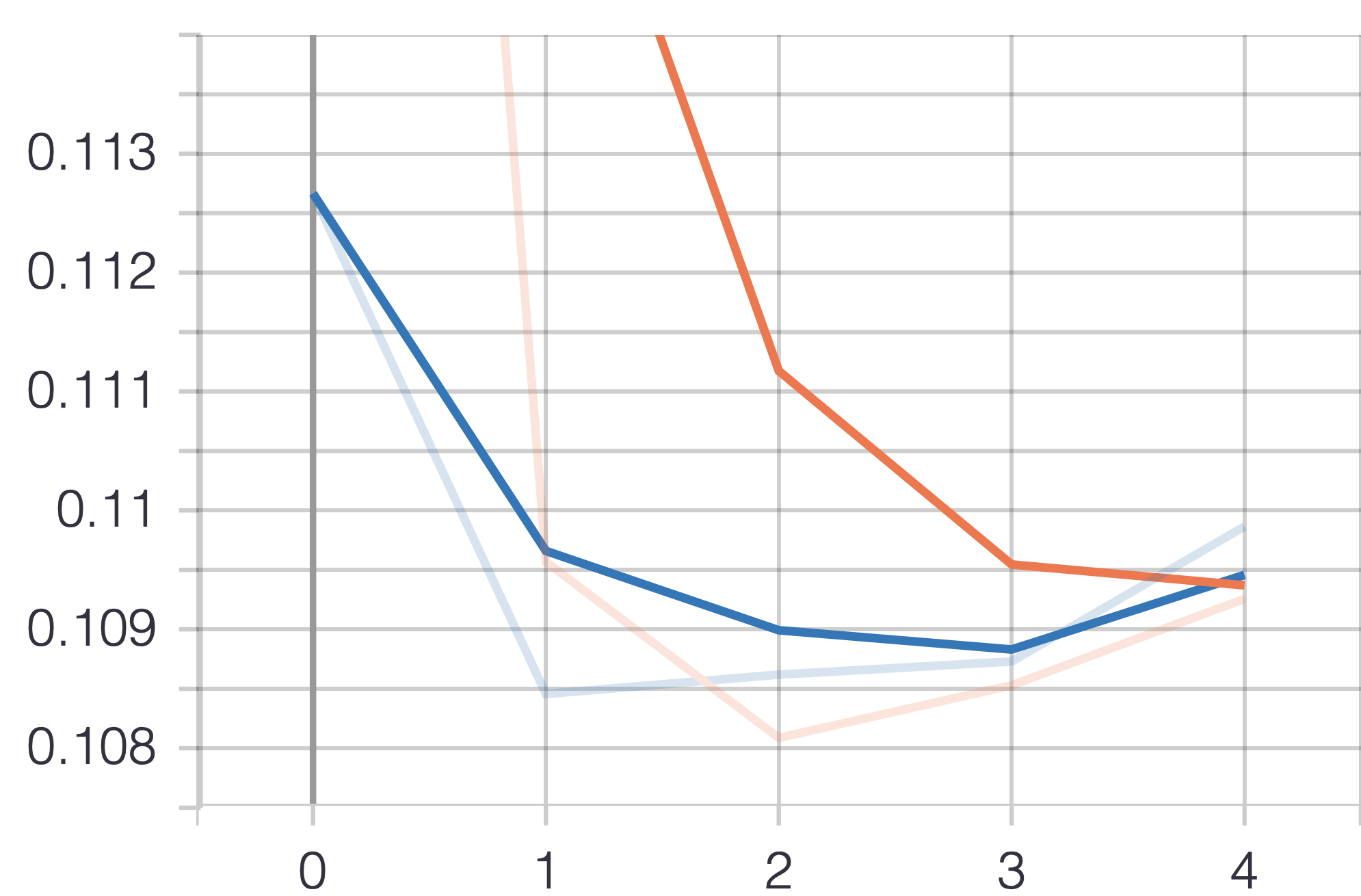}
		 \caption[]{}
		 \label{fig:img_8_2}
	\end{subfigure}
	\caption[]{(a) Training learning curve in terms of batch number, and (b) based on epoch number. Orange and Blues denotes train and validation curves, respectively.}
	\label{fig:img_8}
\end{figure}
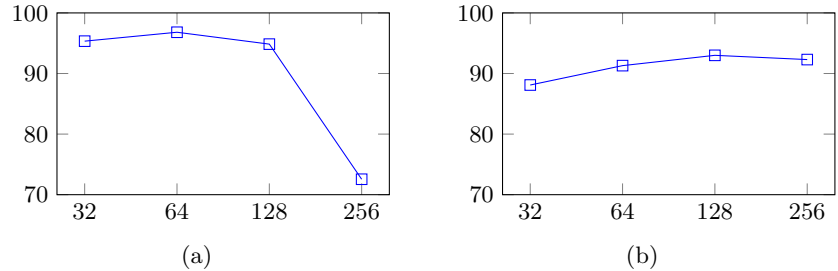
\begin{figure}[t]
	\captionsetup[sub]{justification=centering}
	\centering
	\begin{subfigure}[t]{0.45\textwidth}
		\centering
		\begin{tikzpicture}
			\begin{axis}[
				ymin=70, 
				ymax=100, 
				symbolic x coords={32, 64, 128, 256}, 
				height=4cm, 
				width=6cm]
				\addplot[color=blue, mark=square] coordinates {
					(32, 95.34)(64, 96.81)(128, 94.85)(256, 72.55)};
			\end{axis}
		\end{tikzpicture}
		\caption{}
		\label{fig:img_9_1}
	\end{subfigure}
	\quad
	\begin{subfigure}[t]{0.45\textwidth}
	\centering
	\begin{tikzpicture}
		\begin{axis}[
			ymin=70, 
			ymax=100, 
			symbolic x coords={32, 64, 128, 256}, 
			height=4cm, 
			width=6cm]
			\addplot[color=blue, mark=square] coordinates {
				(32, 88.1)(64, 91.3)(128, 93.0)(256, 92.3)};
			\end{axis}
		\end{tikzpicture}
		\caption{}
		\label{fig:img_9_2}
	\end{subfigure}
	\caption[]{Comparison of effect of number of components in PCA in terms of mAP on the ICDAR 2013 (a) train and (b) test subsets.}
	\label{fig:img_9}
\end{figure}
\begin{table*}[!t]
    \caption{Comparison of evaluation with initial rankings against re-ranking with different values of $k$ on ICDAR 2013 test set.\label{tab:table_1}}
	\centering
	{
		\begin{tabular*}{\textwidth}{@{\extracolsep{\fill}}lcccc}
			\hline
			& Top-1 & Hard-2 & Hard-3 & mAP\\
			\hline
			$\text{ResNet-20+NetVLAD+NN}$ & 98.60 & 84.10 & 65.60 & 93.01\\
			$\text{ResNet-20+NetVLAD+krNN-QE}_{k=1}$ & \textbf{98.70} & 90.30 & 86.70 & 96.48\\
			$\text{ResNet-20+NetVLAD+krNN-QE}_{k=2}$ & 97.90 & 90.50 & \textbf{86.80} & \textbf{96.58}\\
			$\text{ResNet-20+NetVLAD+krNN-QE}_{k=3}$ & 96.40 & \textbf{91.00} & 83.50 & 96.36\\
			\hline
		\end{tabular*}
	}{}
\end{table*}

\subsubsection{Visualization of Embeddings}\label{sub_sub_sec:visualization_of_embeddings}
\begin{figure}
    \centering
    \includegraphics[scale=0.25]{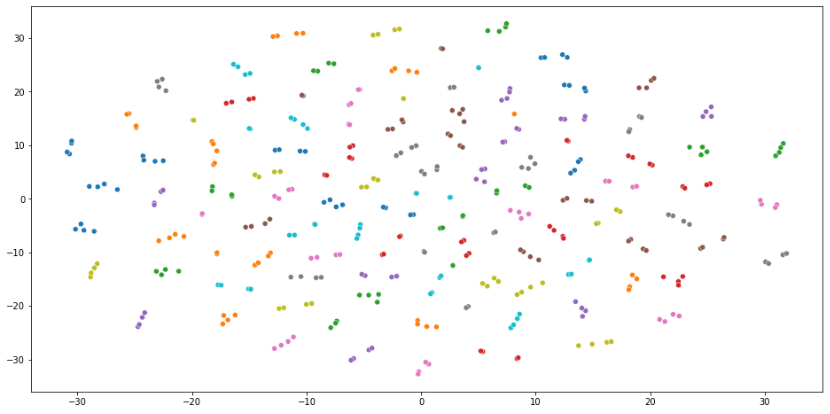}
    \caption{t-SNE plot of the ICDAR 2013 train set global descriptors. There are 100 colors each representing document images of unique writer.}
    \label{fig:img_10}
\end{figure}

\begin{figure}
    \centering
    \includegraphics[scale=0.25]{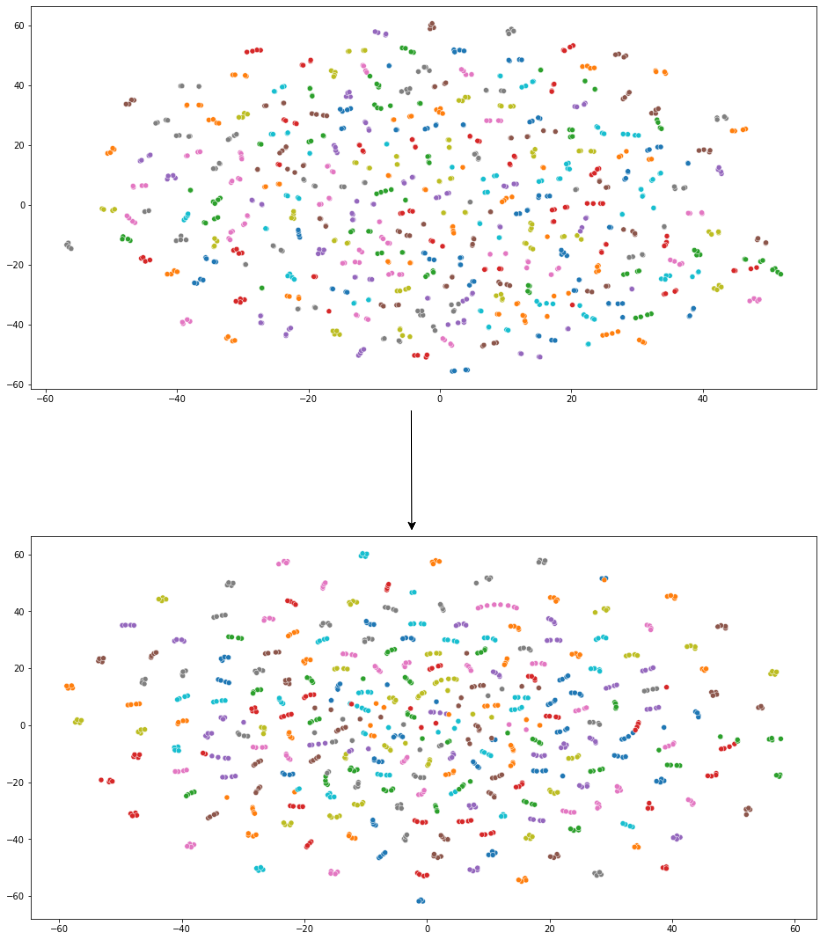}
    \caption{t-SNE plot of ICDAR 2013 test set global descriptors before (above Figure) and after re-ranking (below Figure) with $k = 2$.}
    \label{fig:img_11}
\end{figure}

t-SNE \cite{maaten2008visualizing} is an unsupervised, non-linear technique used mostly for exploration and visualization of high-dimensional data. We perform t-SNE on the ICDAR 2013 train subset to get an intuition of how the computed and dimensionality reduced global descriptors are arranged in the $128$-$D$ space. The t-SNE plot of the embeddings space is shown in Figure \ref{fig:img_10}. The plot shows that the embeddings learned by the proposed pipeline probably have very well discriminative properties as all of the four global descriptors representing each writer's document images approximately lie near each other in the projected $2$-$D$ space then forming a cluster. In other words, the nearest neighbors of each global image descriptor are likely to form the same writer in the original space. On the other hand, Figure \ref{fig:img_11} depicts 1000 global image descriptors of dimension $128$ of the ICDAR 2013 test set before and after the re-ranking part. As the Figure illustrates, the discriminability of clusters is increased and the clusters belonging to each writer are more distinguishable from each other compared to without re-ranking employed.

\subsubsection{Comparison with State of the Art}\label{sub_sub_sec:comparison_with_state_of_the_art}
To compare our results to the state-of-the-art writer recognition pipelines, we deliberate on the three contemporary datasets (ICDAR 2013, CVL, and KHATT). Results are given in terms of Top-1, Hard-2, Hard-3, and mAP evaluation metrics. The Top-1 metric demonstrates the probability of first retrieved items belonging to the same writer as the query. While Hard-2 and Hard-3 are the probabilities representing that all the first two and first three retrieved documents stem from the same writer as the query's, respectively. In most scenarios, the mAP criterion is a more proper complement since it considers the ranking of all the documents. The ResNet-20 convolutional neural network with the NetVLAD layer is employed to extract local descriptors and their corresponding VLAD embeddings. Once embeddings are computed, generalized max-pooling is used to form a global descriptor for each document image. $\ell_{2}$-normalization, SSR, and PCA are subsequently applied to the global descriptors. Note that in all experiments we trained ResNet-20+NetVLAD on the ICDAR 2013 train subset, and used the trained network to evaluate each dataset's test set. In the following, we first summarize our results on these datasets, and secondly, the impression of different parts of the pipeline on the obtained results are investigated. Finally, we also present the overall computational cost of three respective datasets.
\paragraph*{ICDAR 2013.} The document images in this dataset are already in binarized format out-the-box. We train and test and the officially provided train and test subsets of this dataset. Comparison of our results on the ICDAR 2013 dataset is provided in the first section of Table \ref{tab:table_2}. Our pipeline with the proposed re-ranking strategy achieves the overall best result (in terms of mAP) with a 4.2\% difference against the previous best \cite{christlein2018encoding} reported. Another interesting observation is that the Hard-3 criterion is improved by 13.3\%, which is an indicator of the huge benefit that re-ranking brings to the pipeline. This can also be observed on the Hard-2 criterion where around 3.0\% improvement has been achieved. However, the slightly inferior performance obtained in terms of Top-1 which opens the room for further investigation. All in all, comparing results show that our proposed pipeline outperforms all the previous work in this field. Although our pipeline shares some similarities with the pipeline of Christlein et al. \cite{christlein2018encoding}, Our results are far better due to the direct integration of the encoding layer to the ResNet-20, and the re-ranking step.
\paragraph*{CVL.} Since the official train subset of this dataset is rather small (189 samples), we employed the already trained pipeline on the ICDAR 2013 train set and used the official test subset of the CVL dataset containing the subset of writers where each contributed exactly five forms (CVL-283) for evaluation. Also, 
we binarized document images using Otsu's method \cite{otsu1979threshold} to be more similar to the ICDAR 2013 dataset. The obtained results along with the state-of-the-art counterparts provided in the second section of Table \ref{tab:table_2}. Our proposed approach with re-ranking sets new standards by showing superior performance in terms of Hard-3 and mAP and can compete with the state-of-the-art, but slightly behind in terms of Top-1 and Hard-2 metrics.
\paragraph*{KHATT.} The last dataset that we were experimented on is the KHATT dataset. Likewise the CVL dataset, document images are binarized with the Otsu method. The respective results are depicted in the third section of Table \ref{tab:table_2}. We believe that our evaluation strategy on this dataset is slightly different from the state-of-the-art, as they (probably) used the official train set of the respective dataset. Though likewise the CVL dataset, we used the trained network on the ICDAR 2013 officially provided train split and the officially provided test split (600 images from 150 writers) of the respective dataset served as our test set. It is quite interesting that despite training on a dataset containing English and Greek documents, the results are quite promising and comparable to the state-of-the-art.
\begin{table*}[t]
    \caption{Results and comparison with state-of-the-art on the evaluation datasets. Aricles are sorted by their publication date.\label{tab:table_2}}
	\centering
	{
		\begin{tabular*}{\textwidth}{@{\extracolsep{\fill}}llcccc}
			\hline
			Dataset & Method & Top-1 & Hard-2 & Hard-3 & mAP\\
			\hline
			\multirow{7}{1.5em}{ICDAR 2013} & $\text{RootSIFT+SV+NN}$ \cite{christlein2014writer} & 97.1 & 42.8 & 23.8 & 67.1\\
			& $\text{Contour-Zernike+VLAD+NN}$ \cite{christlein2015writer} & 99.4 & 81.0 & 61.8 & 88.0\\
			& $\text{LeNet-5+SV+NN}$ \cite{christlein2015offline} & 98.9 & 83.2 & 61.3 & 88.6\\
			& $\text{CaffeNet+NN}$ \cite{fiel2015writer} & 88.5 & 63.2 & 36.5 & -\\
			& $\text{RootSIFT+SV+ESVM}$ \cite{christlein2017writer} & 99.7 & 84.8 & 63.5 & 89.4\\
			& $\text{ResNet-20+VLAD+ESVM}$ \cite{christlein2018encoding} & \textbf{99.6} & 89.8 & 77.0 & 93.2\\
			& $\text{ResNet-20+NetVLAD+krNN-QE}_{k=2}$ & 97.4 & \textbf{92.8} & \textbf{90.3} & \textbf{97.41}\\
		    \hline
		    \multirow{6}{1.5em}{CVL} & $\text{RootSIFT+SV+NN}$ \cite{christlein2014writer} & 99.2 & 98.1 & 95.8 & 97.1\\
		    & $\text{Contour-Zernike+VLAD+NN}$ \cite{christlein2015writer} & 99.4 & 98.9 & 97.4 & 97.9\\
		    & $\text{LeNet-5+SV+NN}$ \cite{christlein2015offline} & 99.4 & 98.8 & 97.3 & 97.8\\
		    & $\text{CaffeNet+NN}$ \cite{fiel2015writer} & 88.5 & 63.2 & 36.5 & -\\
		    & $\text{RootSIFT+SV+ESVM}$ \cite{christlein2017writer} & 99.2 & 98.4 & 97.1 & 98.0\\
		    & $\text{ResNet-20+VLAD+ESVM}$ \cite{christlein2018encoding} & \textbf{99.5} & \textbf{99.0} & 97.7 & 98.4\\
		    & $\text{ResNet-20+NetVLAD+krNN-QE}_{k=2}$ & 99.2 & 98.9 & \textbf{98.0} & \textbf{98.6}\\
			\hline
			\multirow{5}{1.5em}{KHATT} & $\text{RootSIFT+SV+NN}$ \cite{christlein2014writer} & 99.3 & 96.8 & 94.5 & 97.5\\
			& $\text{Contour-Zernike+VLAD+NN}$ \cite{christlein2015writer} & 99.4 & \textbf{98.9} & \textbf{97.4} & 97.9\\
			& $\text{RootSIFT+SV+ESVM}$ \cite{christlein2017writer} & 99.5 & 96.5 & 92.5 & 97.2\\
			& $\text{ResNet-20+VLAD+ESVM}$ \cite{christlein2018encoding} & \textbf{99.6} & 97.6 & 94.5 & \textbf{98.0}\\
			& $\text{ResNet-20+NetVLAD+krNN-QE}_{k=2}$ & 98.6 & 95.3 & 93.0 & 97.7\\
			\hline
		\end{tabular*}
	}{}
\end{table*}

\subsubsection{Influence of Different Parts of the Pipeline}\label{sub_sub_sec:influence_of_different_parts_of_the_pipeline} 
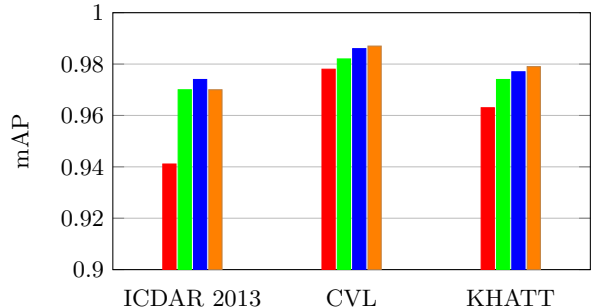
\begin{figure}[ht]
    \centering
    \begin{tikzpicture}
    \begin{axis}[
        width = 0.65*\textwidth, 
        height = 5cm, 
        major x tick style = transparent,
        ybar=2*\pgflinewidth,
        bar width=5pt,
        ymajorgrids = true, 
        ylabel = {mAP},
        symbolic x coords={ICDAR 2013, CVL, KHATT},
        xtick = data,
        scaled y ticks = false,
        enlarge x limits=0.25, 
        ymin=0.9, ymax=1
    ]
        \addplot[style={red, fill=red, mark=none}]
            coordinates {(ICDAR 2013, 0.9411) (CVL, 0.978) (KHATT, 0.963)};
        \addplot[style={green, fill=green, mark=none}]
             coordinates {(ICDAR 2013, 0.970) (CVL, 0.982) (KHATT, 0.974)};
        \addplot[style={blue, fill=blue, mark=none}]
             coordinates {(ICDAR 2013, 0.974) (CVL, 0.986) (KHATT, 0.977)};
        \addplot[style={brown, fill=orange, mark=none}]
             coordinates {(ICDAR 2013, 0.970) (CVL, 0.987) (KHATT, 0.979)};
        \legend{}
    \end{axis}
    \end{tikzpicture}
    \caption{Comparison of using $\text{NN}$ or $\text{krNN-QE}_{k}$ to rank documents. Red: $\text{NN}$, Greean: $\text{krNN-QE}_{k=1}$, Blue: $\text{krNN-QE}_{k=2}$, and Orange: $\text{krNN-QE}_{k=3}$.}
    \label{fig:img_12}
\end{figure}

Most of the writer recognition pipelines are made out of three major parts: (1) Feature Extraction, (2) Encoding, and (3) ranking strategy. As we have demonstrated in Table \ref{tab:table_3} we have specified each method name by these three parts.\\
\paragraph*{Feature Extraction and Encoding.} 
Feature Extraction and Encoding steps are two inseparable parts of the writer recognition pipelines. In contrast to the numerous hand-crafted feature methods such as Countor-Zernike and SIFT descriptors, we use features learned by a convolutional neural network. relying on the neural networks brings the advantage of being data-driven. As data progresses through the CNN layers, a higher level of abstraction is reached automatically by the data. Also, the encoding step is required for the pipeline to computing a single representation for each document image from its many feature descriptors. There are a few encoding methods such as GMM Supervectors, VLAD, I-Vectors, etc. used in the literature where specifically VLAD encoding results were quite promising. However, none of them are deep-learning-based and we believe that constructing a unified neural network consisting of both the feature extraction and encoding step as one integrated step would bring many advantages to the system. One of the first deep-learning-based approaches is implemented by Fiel et al. \cite{fiel2015writer} employed the CaffeNet which is part of the "Caffe Deep Learning Framework". However, they trained their convolutional neural network on the word images of the IAM dataset and extracted the features of the penultimate fully connected layer for evaluation. But they did not use any specific encoding method and this might a reason that their approach is far behind the state-of-the-art methods using an encoding step. Christlein et al. \cite{christlein2018encoding} already used ResNet-20 as a feature descriptor, However, we also integrate VLAD encoding step into the neural network by employing the NetVLAD layer in the head of ResNet-20. In this manner, We have one unified neural network to extract embeddings from each document image, To combine Feature extraction and encoding in one part. Table \ref{tab:table_3} compares our unified feature extraction and encoding step with others in the literature.
\begin{table}[h]
    \caption{Effect of feature extraction and encoding step of the pipeline on the ICDAR13 test set.\label{tab:table_3}}
    \centering
    {
        \begin{tabular*}{20pc}{@{\extracolsep{\fill}}lr}
            \hline
            Method & mAP\\
            \hline
            RootSIFT+SV & 67.1\\
            Contour-Zernike+VLAD & 88.0\\
            LeNet-5+SV & 88.6\\
            ResNet-20+VLAD & 93.2\\
            ResNet-20+NetVLAD & 94.1\\
            \hline
        \end{tabular*}
    }{}
\end{table}
\paragraph*{Ranking Strategy.} Routinely, in the leave-on-image-out cross-validation, the ranking of other documents are computed based on cosine similarity with the nearest neighbor approach. However, performance could be leveraged by the knowledge provided in this initial ranking in order to refine them. While it is quite common in image retrieval methods to use re-ranking strategies, It has not brought much attention to it in the field of writer identification and writer retrieval. To the best of our knowledge, Jordan et al. in \cite{jordan2020re} are the only ones who employed this scenario. However, their approach differs from ours since they used the query expansion to add more positive examples before evaluation using ESVMS. To demonstrate the usefulness of our approach, in Table \ref{fig:img_12} we have provided how the mAP criterion changes according to the ranking strategy combined with different parameter values for $k$ on the evaluated datasets. It can be out-turned from the bar plot that our re-ranking strategy brings huge benefits to the pipeline improving the mAP criterion.

\subsubsection{Computational Cost}\label{sub_sub_sec:computational_cost} 
Our ResNet-20 with the NetVLAD layer consists of nearly 285,000 trainable parameters. In our experiments, training time takes about 75 minutes in a Google Colab's GPU Hardware accelerator of type NVIDIA Tesla P100-PCIE-16GB. Also, Table \ref{tab:table_4} summarizes the computation time of the global descriptors of evaluated datasets. Note that, the computation involves patch extraction, embeddings extraction from the pre-trained model, and ridge regression optimization steps.

\begin{table}[h]
    \caption{Comparison of global descriptors computation time.\label{tab:table_4}}
    \centering
    {
        \begin{tabular*}{20pc}{@{\extracolsep{\fill}}lcc}
            \hline
            Dataset & Images & Elapsed time\\
            \hline
            ICDAR 2013 (Train) & 400 & 2935s\\
            ICDAR 2013 (Test) & 1000 & 10255s\\
            CVL (Test) & 1415 & 14524s\\
            KHATT (Test) & 600 & 5988s\\
            \hline
        \end{tabular*}
    }{}
\end{table}

\section{Conclusion}\label{sec:5_conclusion}
In this work, we have presented a 1) novel pipeline consisting of a convolutional neural network followed by the NetVLAD layer to extract local descriptors and their corresponding VLAD embeddings in an end-to-end manner and 2) re-ranking strategy based on query expansion and $k$-reciprocal nearest neighbors to improve initial rankings.\\
\indent Our results demonstrate improvements and set new standards on both ICDAR13 and CVL datasets. Having in mind that we evaluated KHATT data using the pre-trained network on ICDAR 2013 dataset, The results on this dataset is quite promising as well.  However, there is still room for improvement and optimization of the proposed pipeline in various directions. Since we did not preprocess image patches for input to the network, the preprocessing step could be investigated in more detail. Also, deep learning-based approaches other than NetVLAD such as DeepTen \cite{zhang2017deep} may worth investigating. On the other hand, we have used the NetVLAD layer to extract embeddings but employing it to directly learn global image descriptors could also be beneficial. Finally, historical data are getting more and more attention in recent years, so for future works, the application of the proposed pipeline on historical data must be researched.


\section{Conflict of Interest}\label{sec:7_conflict_of_interest}
The authors have no conflict of interest to declare.

\end{document}